\documentclass{article} 
\usepackage{iclr2027_conference,times}

\usepackage{amsmath,amsfonts,bm}

\def\eqref#1{equation~\ref{#1}}

\def\1{\bm{1}}

\DeclareMathAlphabet{\mathsfit}{\encodingdefault}{\sfdefault}{m}{sl}
\SetMathAlphabet{\mathsfit}{bold}{\encodingdefault}{\sfdefault}{bx}{n}

\usepackage{hyperref}
\usepackage{url}
\usepackage{algorithm}
\usepackage{algorithmic}
\usepackage{amsfonts}
\usepackage{multirow}
\usepackage{booktabs}   
\usepackage{makecell}   
\usepackage{graphicx}   
\usepackage{natbib}     
\usepackage[table]{xcolor}
\usepackage{newfloat}
\usepackage[dvipsnames]{xcolor}

\usepackage{enumitem}

\title{LexReward: A Taxonomy-Driven Reward Framework for Legal Language Models}

\author{
{\bfseries
Yida Cai$^{1,3}$\thanks{Equal contribution. Research conducted during Yida Cai’s internship at Tsinghua University.},~
Xin Dai$^{2}$\footnotemark[1],~
Bingxiang He$^{3}$,~
Huiyuan Xie$^{3}$\thanks{Corresponding author.},~
Yuxiao Ye$^{3}$},~
Zhenghao Liu$^{2}$, \\
{\bfseries
~Yang Bai$^{1}$,
Zhiyuan Liu$^{3}$} \\
$^{1}$Peking University
$^{2}$Northeastern University
$^{3}$Tsinghua University
\\
\small{~caiyida26@stu.pku.edu.cn, 20216401@stu.neu.edu.cn, xieh@tsinghua.edu.cn}
}

\iclrfinalcopy 
\begin{document}

\maketitle

\begin{abstract}
Legal language models require reward signals that capture not only answer correctness but also the multidimensional quality of legal responses. Existing reward methods, however, often rely on coarse-grained holistic judgments, providing limited domain specificity and interpretability. We introduce \textbf{LexReward}, a taxonomy-driven framework for legal reward modeling. \textbf{LexReward} characterizes legal response quality along three complementary dimensions: \textit{Style}, covering lexical and syntactic quality; \textit{Element}, assessing legal subjects, facts, statutes, and decisions; and \textit{Chain}, evaluating the order, completeness, correctness, and non-redundancy of legal reasoning. For each dimension, we develop rubrics that specify evaluation criteria and quality levels. 
The resulting rewards are used to construct pairwise preference data for Direct Preference Optimization (DPO) and reward-model training. Experiments show that the rubric-based rewards reliably distinguish legal responses of different quality and that DPO training on the preference data improves performance across all three dimensions. The learned reward models, \textbf{LexRM}, also support effective downstream optimization: each dimension-specific reward model improves policy performance in its corresponding dimension through reinforcement learning, without requiring reference answers at reward time. Dimension-wise analyses further support the effectiveness of the proposed taxonomy and reward construction.\footnote{Data and code are available at \url{https://github.com/thunlp/LexReward}.} 

\end{abstract}

\section{Introduction}

Legal large language models have demonstrated growing potential across a wide range of legal generation tasks, such as legal question answering and legal reasoning~\citep{lawbench,xie2026lexchain,yao-etal-2025-elevating,li2024lexevalcomprehensivechineselegal}.
Nevertheless, producing a high-quality legal response requires considerably more than arriving at the correct final answer. 
A legally sound response should identify the relevant subjects and facts, accurately invoke applicable statutes, construct a complete and coherent reasoning chain, and express its conclusions in precise and objective legal language. 
Models may reach a correct conclusion through incomplete reasoning, omit legally decisive facts, cite an inappropriate statutory basis, or produce fluent but legally unreliable explanations. These failures cannot be adequately captured by final-answer accuracy alone.

As reinforcement learning (RL) becomes increasingly important for improving model reasoning, reward design plays a central role in specifying which aspects of legal response quality models are encouraged to improve.
Existing reward approaches in legal AI often rely on general-purpose reward models or reward functions that assess only selected aspects of legal response quality, such as judgment-outcome accuracy~\citep{cai2025unilaw,zhang2025legal}. Although useful in some settings, these approaches may overlook other aspects of a response, making it difficult to determine whether improvements reflect better legal response quality or superficial features such as response length and fluency. 
More fundamentally, without a structured definition of legal response quality, it remains unclear what a legal reward should assess.

\begin{figure}[t]
  \centering
  \includegraphics[width=\columnwidth]{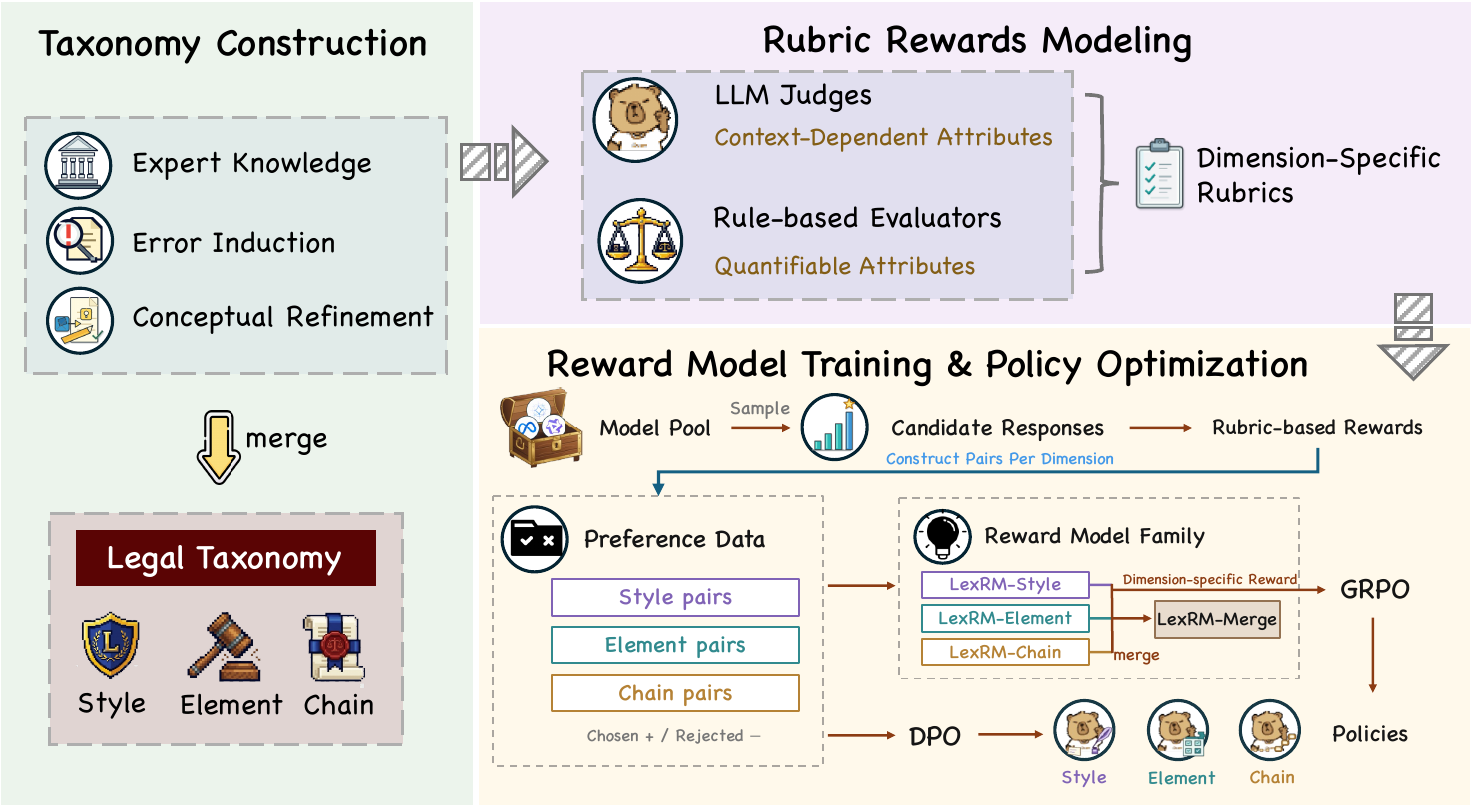}
  \caption{Overview of the LexReward framework. LexReward first constructs a legal taxonomy through expert knowledge, error induction, and conceptual refinement. It then
operationalizes the taxonomy into dimension-specific rubric-based rewards, which construct preference
data for training a reward model and guide reinforcement learning.}
  \label{fig:lexreward-main}
\end{figure}

To clarify which aspects of legal response quality should be rewarded, as shown in Fig.~\ref{fig:lexreward-main}, we introduce \textbf{LexReward}, a taxonomy-driven reward framework for legal language models.
Drawing on legal experts' domain knowledge, we first develop a taxonomy of legal response quality with three complementary dimensions:
\textit{Style}, which captures lexical and syntactic properties of legal writing; \textit{Element}, which assesses the identification and treatment of legal subjects, facts, statutes, and decisions; and \textit{Chain}, which evaluates the order, completeness, correctness, and non-redundancy of legal reasoning. 
Legal experts further refine the taxonomy through an empirical analysis comparing existing models' responses with reference answers, yielding fine-grained criteria for each dimension.
We design a scoring rubric for each fine-grained criterion and use either rule-based or LLM-as-a-judge evaluators, depending on the context. Experiments show that the resulting rewards reliably distinguish responses of different quality along their corresponding dimensions.

\begin{figure}[t]
  \centering
  \includegraphics[width=\textwidth]{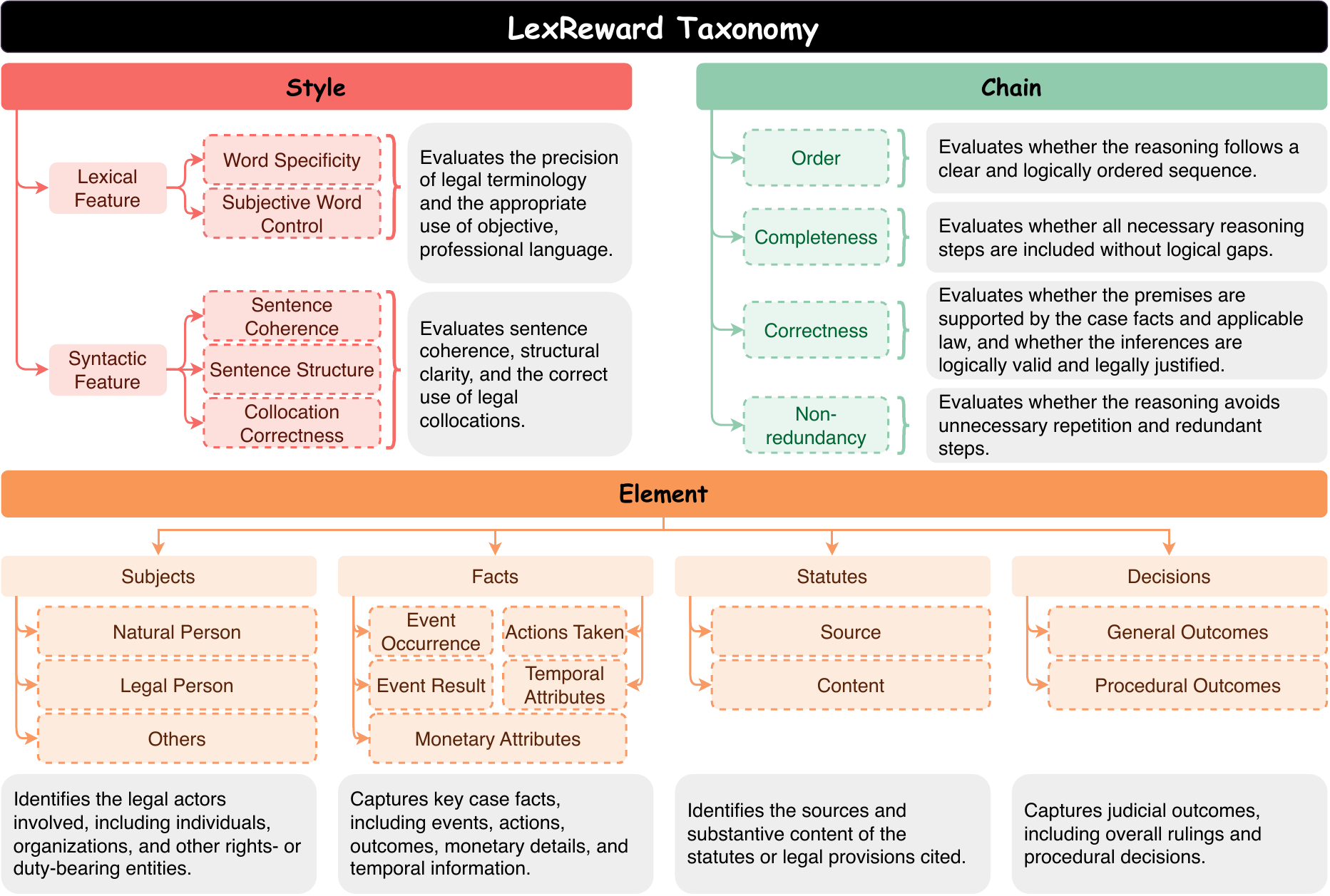}
  \caption{
  The LexReward taxonomy and its rubric-oriented operationalization. Legal response quality is decomposed into three complementary dimensions: \textit{Style} (red), \textit{Element} (orange), and \textit{Chain} (green). Colored boxes show the dimensions and their constituent criteria, while gray boxes summarize the corresponding evaluation objectives. This structured decomposition supports the development of fine-grained rubrics for legal reward modeling.}
  \label{fig:lexreward-framework}
\end{figure}

To provide dense reward signals and evaluate the practical utility of \textbf{LexReward}, we score diverse responses from a pool of models using rubric-based rewards and construct pairwise preference data both to evaluate their utility for Direct Preference Optimization (DPO), and to train \textbf{LexRM}, a family of Chinese legal reward models. To the best of our knowledge, \textbf{LexRM} is the first collection of reward models developed for the Chinese legal context. 
We evaluate \textbf{LexRM} through Test-Time Scaling (TTS), using it to select the highest-scoring response from candidates generated by multiple models and comparing against random selection from the same candidate pool.
We further assess its effectiveness for Group Relative Policy Optimization (GRPO).
Our experiments show that DPO improves generation performance across all three dimensions, while \textbf{LexRM}-guided selection outperforms random selection. When used for GRPO, \textbf{LexRM} also yields gains across all dimensions and outperforms rule-based outcome rewards~\citep{legaldelta, cai2025unilaw, zhang2025legal}.

Together, these results establish \textbf{LexReward} as a framework for defining and rewarding legal response quality, enabling legal domain knowledge to guide both reward construction and model optimization.
Our main contributions are as follows:
\begin{itemize}[leftmargin=*,topsep=1pt,itemsep=1pt]
    \item We introduce a structured taxonomy of legal response quality that decomposes the requirements of a high-quality legal response into three complementary dimensions (\textit{Style}, \textit{Element}, and \textit{Chain}) and fine-grained criteria, providing a principled foundation for legal reward modeling.

    \item We develop interpretable, rubric-based rewards for each fine-grained criterion in the taxonomy using rule-based and LLM-based evaluation, and demonstrate that they reliably distinguish response quality along their respective dimensions.

    \item We construct preference data using the rubric-based rewards and demonstrate that DPO training on these data improves legal response generation.

    \item We train \textbf{LexRM}, to the best of our knowledge the first family of Chinese legal reward models, on rubric-derived preference data and demonstrate its effectiveness in Test-Time Scaling (TTS).

    \item We further integrate the learned reward models into a Group Relative Policy Optimization (GRPO) pipeline and demonstrate that their supervision improves the generation quality of the policy.
\end{itemize}

\section{Related Work}
\label{sec:related}

\subsection{Legal Language Models and Domain-Specific Evaluation}

Large language models (LLMs) have been increasingly adapted to the legal domain, with prior work improving their legal knowledge and reasoning abilities through domain-specific training and alignment methods~\citep{chalkidis2020legalbert,legalduet,legaldelta}. Meanwhile, benchmarks such as LegalBench~\citep{legalbench} and LawBench~\citep{lawbench} provide standardized testbeds for evaluating legal knowledge, rule understanding, and reasoning ability across different legal tasks~\citep{zhong2020jecqa}.
However, existing legal evaluations often emphasize task-level performance, which does not fully capture the multidimensional quality of a legal response. A reliable legal response should also exhibit appropriate legal expression, sufficient coverage of relevant legal elements, faithful grounding in applicable statutes, and coherent reasoning. This motivates a structured reward framework that decomposes legal response quality into interpretable dimensions and translates them into training and optimization signals.

\subsection{Rubric-based Reward Modeling}
Reward modeling learns to assess response quality for candidate selection and policy optimization. Early approaches learn scalar rewards from holistic human preferences~\citep{ouyang2022training}, while subsequent work introduces finer-grained supervision over error types, text segments, and multiple quality objectives~\citep{wu2023finegrained,wang2024interpretable}. Complementing this decomposition, rubric-based evaluation~\citep{kim2024prometheus} makes judgment criteria and quality levels explicit, enabling evaluators to assess responses against specified standards. In the legal domain, existing work~\citep{chen2026scprm} learns process rewards for criminal-law knowledge-graph reasoning, but focuses on a specific task. Building on the broader shift from holistic preferences toward structured quality supervision, \textbf{LexReward} systematically defines reward dimensions and evaluation criteria through a legal quality taxonomy, then operationalizes it into rubrics and cross-task preference data, connecting domain-specific quality definitions with reward model training and reinforcement learning.

\section{The LexReward Taxonomy}
\paragraph{Taxonomy construction.} Legal response quality encompasses multiple requirements that a single holistic criterion leaves implicit. We therefore construct a taxonomy that decomposes legal response quality into explicit, assessable dimensions. With the assistance of legal experts, we combine top-down specification with bottom-up error analysis. In the top-down component, legal experts draw on their domain knowledge to identify the core requirements of a high-quality legal response~\citep{goodrich1990legal,osbeck2011good,maley2014language}. In the bottom-up component, they examine responses generated by multiple language models~\citep{qwen3technicalreport,dubey2024llama} across legal AI benchmarks~\citep{lawbench,ma2026clase,li2024lexevalcomprehensivechineselegal}, comparing them with reference answers to identify errors and unmet quality requirements. The requirements identified through both components are then integrated and refined into a unified taxonomy.

As illustrated in Fig.~\ref{fig:lexreward-framework}, the resulting taxonomy comprises three complementary dimensions: \textit{Style}, \textit{Element}, and \textit{Chain}. These dimensions characterize how legal content is expressed, what legally relevant information is included, and how that information is connected to support a conclusion. Each dimension is further decomposed into finer-grained criteria for rubric-based assessment.

\textbf{Style} assesses the linguistic quality of legal responses through \textit{lexical features} and \textit{syntactic features}. \textit{Lexical features} capture the precise use of legal terminology (\textit{word specificity}) and the use of objective language without unwarranted subjective judgments (\textit{subjective word control}). \textit{Syntactic features} capture cohesion across sentences (\textit{sentence cohesion}), clarity of sentence structure (\textit{sentence structure}), and appropriate combinations of words in legal expressions (\textit{collocation correctness}). Together, these criteria assess the precision, objectivity, and clarity of legal expressions.

\textbf{Element} assesses whether a response includes the legally relevant information needed to address a legal problem. It comprises four categories: \textit{subjects}, \textit{facts}, \textit{statutes}, and \textit{decisions}, covering the relevant entities, case circumstances, statutory provisions, and legal conclusions, respectively.

\textbf{Chain} assesses the reasoning that connects case information to a legal conclusion. It comprises four categories: \textit{order}, \textit{completeness}, \textit{correctness}, and \textit{non-redundancy}. These criteria assess whether the reasoning steps are logically arranged, sufficiently developed, legally valid, and free from unnecessary repetition. Whereas \textit{Element} assesses coverage of relevant information, \textit{Chain} assesses the inferential connections among that information.

Together, the three dimensions organize legal response quality in terms of stylistic expression, substantive content, and reasoning. The taxonomy provides a structured basis for translating expert-defined quality requirements into explicit rubric criteria, which support fine-grained assessment and rubric-based reward modeling.

\section{Taxonomy-Driven Rewards}

\subsection{Rubric-based Reward Operationalization}

We operationalize the fine-grained criteria in our taxonomy using two complementary scoring strategies, selected according to whether evaluating a criterion requires the input context. For \textit{context-independent} criteria, the score can be computed from intrinsic properties of the generated response. We therefore use rule-based reward functions calibrated against a corpus of high-quality legal texts, providing deterministic and interpretable scores. For \textit{context-dependent} criteria, evaluation requires determining whether the response appropriately addresses the legal problem specified by the input. We use an LLM-as-a-judge evaluator for these criteria, leveraging its ability to assess the relationship between the input and the generated response under a criterion-specific rubric.

For a context-independent criterion \(k\), we define the reward as:
\begin{equation}
r_k(y)
=
\operatorname{Eval}^{\mathrm{rule}}_k
\left(y; \theta_k, \mathcal{C}\right),
\end{equation}
where \(y\) denotes the generated response, \(\theta_k\) denotes the criterion-specific scoring parameters, and \(\mathcal{C}\) is an optional corpus of high-quality legal texts used to calibrate the evaluator.

For a context-dependent criterion \(k\), we define the reward as:
\begin{equation}
r_k(x,y)
=
\operatorname{Eval}^{\mathrm{LLM}}_k
\left(x,y; \rho_k\right),
\end{equation}
where \(x\) denotes the input context, \(y\) denotes the generated response, and \(\rho_k\) specifies the scoring rubric supplied to the LLM judge.

We assign an evaluation strategy to each taxonomy criterion based on its dependence on the input context. \textit{Style} primarily concerns intrinsic linguistic properties of the response and is therefore evaluated using a rule-based evaluator. \textit{Element} requires assessing whether the response identifies and treats the legally relevant information in the input and is therefore evaluated using an LLM judge. The evaluation of \textit{Chain} is task-dependent. When a task provides an explicit reasoning structure, we use rule-based functions to assess conformity to the prescribed steps and order. When the appropriate reasoning path must be inferred from the legal context, we instead use an LLM judge. We next describe the rubric and scoring procedure for each fine-grained criterion:

\textbf{Style.} The \textit{Style} reward measures how closely a response conforms to the lexical and syntactic patterns of authentic Chinese judicial documents. We evaluate five attributes:
$\mathcal{K}_{\mathrm{Style}} =
\{\mathrm{ws}, \mathrm{sw}, \mathrm{coh}, \mathrm{str}, \mathrm{col}\}$,
corresponding to \textit{word specificity}, \textit{subjective word control}, \textit{sentence cohesion}, \textit{sentence structure}, and \textit{collocation correctness}, respectively.

For each attribute $k$, we extract a feature representation $f_k(y)$ from response $y$ and measure its discrepancy from the corresponding reference representation $f_k^{\mathrm{ref}}$, estimated from a corpus of authentic judicial documents:
\begin{equation}
d_k(y)
=
D_k\!\left(f_k(y), f_k^{\mathrm{ref}}\right),
\qquad k \in \mathcal{K}_{\mathrm{Style}},
\end{equation}
where $D_k$ is a non-negative discrepancy measure appropriate to the feature type. \textit{Word specificity} and \textit{subjective word control} use KL divergence to compare legal-term and sentiment-word distributions, respectively. \textit{Sentence cohesion} uses the absolute difference in conjunction frequency. \textit{Sentence structure} and \textit{collocation correctness} use Euclidean distance to compare sentence-length statistics and $3$- to $6$-gram coverage vectors, respectively.

To account for scale differences, we divide each discrepancy by its mean over calibration corpus $\mathcal{C}$:
\begin{equation}
s_k
=
\frac{1}{|\mathcal{C}|}
\sum_{y' \in \mathcal{C}} d_k(y'),
\qquad
\widetilde{d}_k(y)
=
\frac{d_k(y)}{s_k+\epsilon},
\end{equation}
where $\epsilon>0$ ensures stability. The \textit{Style} reward is the negated mean normalized discrepancy:
\begin{equation}
R_{\mathrm{Style}}(y)
=
-\frac{1}{|\mathcal{K}_{\mathrm{Style}}|}
\sum_{k \in \mathcal{K}_{\mathrm{Style}}}
\widetilde{d}_k(y).
\label{eq:style-reward}
\end{equation}
Higher rewards indicate greater similarity with the reference writing style across the five attributes. Feature extraction, reference estimation, and calibration procedures are detailed in Appendix~\ref{appendix-style}.

\textbf{Element.} The \textit{Element} rubric evaluates whether a legal response matches the task-required legal components. We instantiate this rubric with an LLM-as-a-Judge evaluator, which scores the candidate response along four sub-dimensions: \textit{subjects}, \textit{facts}, \textit{statutes}, and \textit{decisions}. \textit{Subjects} capture legal actors and their roles; \textit{facts} capture case-relevant factual information; \textit{statutes} capture statutory grounds; and \textit{decisions} capture the final legal conclusion. Given a query and a candidate response, the judge assigns each sub-dimension a score from 0 to 1 without access to the reference answer, and the \textit{Element} reward is computed as the average of the four scores. The judge is instructed to assess whether the candidate answer matches the query requirements in each \textit{Element} sub-dimension, considering element accuracy, coverage, specificity, and unsupported or fabricated content. The full prompt is provided in Appendix~\ref{appendix-element}.

\textbf{Chain.} The \textit{Chain} rubric evaluates the structure of legal reasoning along four sub-dimensions: \textit{order}, \textit{completeness}, \textit{correctness}, and \textit{non-redundancy}. 
\textit{Order} assesses the logical sequence of reasoning steps; \textit{completeness} measures coverage of required steps; \textit{correctness} checks whether each step serves an appropriate reasoning function; and \textit{non-redundancy} assesses unnecessary repetition or conflicting content across repeated steps. 
When a task specifies predefined reasoning steps, a rule-based evaluator scores these dimensions against the prescribed step types, required groups, and ordering constraints. Otherwise, an LLM-as-a-Judge evaluates the response along the same dimensions based on the query requirements. The dimension-level scores are aggregated into the final \textit{Chain} reward. Detailed scoring rules and the judge prompt are provided in Appendix~\ref{appendix-chain}.

\subsection{Reward Model Training from Rubric-Guided Preferences}
Rubric-based rewards can be sparse, often assigning identical scores to responses of differing quality. Such ties limit their ability to distinguish between candidate responses and can reduce their effectiveness in downstream applications. 
We therefore construct preference data from rubric-derived scores and train \textbf{LexRM}, a family of Chinese legal reward models, to learn legal preference functions that generalize beyond the discrete distinctions captured by the rubrics.

\textbf{Dimension-specific reward models.} To diversify response sources and reduce reliance on stylistic cues specific to a single generator, we maintain a pool of $n$ models that generate candidate responses for each query $x$. For each quality dimension $d$, we score these candidates using the corresponding rubric and construct preference pairs within the same query. To capture quality differences across the score range, we organize the pairs into three categories: (i) the highest-scoring response versus the nearest lower-scoring response in the high-score range; (ii) a high-scoring response versus a medium-scoring response; and (iii) a medium-scoring response versus a substantially lower-scoring response. These categories provide supervision for both fine-grained distinctions among strong responses and broader differences across quality levels. 

The resulting preference dataset $\mathcal{P}_d$ contains tuples $(x,y^+,y^-)$, where $y^+$ receives a higher rubric score than $y^-$. We train a separate reward model $R_{\theta_d}$ per dimension using the Bradley--Terry objective~\citep{bradley1952rank}:
\begin{equation}
\mathcal{L}_{\mathrm{BT}}^{(d)}
=
-\mathbb{E}_{(x,y^+,y^-)\sim\mathcal{P}_d}
\left[
\log \sigma\!\left(
R_{\theta_d}(x,y^+) - R_{\theta_d}(x,y^-)
\right)
\right],
\end{equation}
where $\sigma$ denotes the sigmoid function. This objective encourages the model to assign higher scalar scores to preferred responses. The resulting dimension-specific reward models support separate assessment of each taxonomy dimension, enabling analysis of where legal response quality improves or remains deficient.
\textbf{Multi-dimensional reward models.} The dimension-specific reward models, one per dimension $d\in\mathcal{D}=\{\textit{element},\textit{style},\textit{chain}\}$, share one pretrained backbone $\theta_0$ and differ only in the preference datasets used for fine-tuning. For each dimension, we represent the parameter update as a task vector $\tau_d=\theta_d-\theta_0$. Across the backbone's weight matrices, which carry virtually all of its parameters, the pairwise cosine similarities between these task vectors are small ($|\cos|\le0.03$ on average, never above $0.14$), with the little overlap that exists confined to the value and output projections of the topmost layers. These low pairwise cosine similarities motivate exploring additive composition of the dimension-specific updates. We form a single multi-dimensional reward model using task arithmetic~\citep{ilharco2023editing}:
\begin{equation}
\theta_{\mathrm{merged}}=\theta_0+\textstyle\sum_{d\in\mathcal{D}}\lambda_d\tau_d,\qquad \lambda_d=1,
\end{equation}
where $\lambda_d$ scales the update contributed by dimension $d$. The merged model is a single reward model that scores all dimensions, removing the need to run three backbones at inference time.



\section{Experiments and Results}

\begin{table*}[t]
\centering
\caption{
Performance of rubric-based rewards and reward models on datasets evaluating three dimensions. Bold denotes the best result in each column.
}
\label{tab:rm_results}

\begin{minipage}{\textwidth}
\footnotesize
\setlength{\tabcolsep}{2pt}
\renewcommand{\arraystretch}{1.12}

\noindent
\begin{tabular*}{\linewidth}
{@{\extracolsep{\fill}}lr@{\hspace{6pt}}lrrrrrr@{}}
\toprule
\multicolumn{2}{c}{\textbf{Style: CLASE}} &
\multicolumn{7}{c}{\textbf{Element: Legal$\Delta$}} \\
\cmidrule(lr){1-2}\cmidrule(l){3-9}
Method & Accuracy (\%) &
Method & SPP-F & CCP & SLP & CAS & CAC & Avg. \\
\midrule
Random & 50.00 &
Five-model average & 53.37 & 46.34 & 28.46 & 35.88 & 84.12 & 49.63 \\
Rubric & 79.00 &
Rubric & \textbf{79.04} & 57.03 & 33.10 & 39.40 & \textbf{92.80} & 60.27 \\
Skywork-Qwen & 78.50 &
Skywork-Qwen & 75.97 & 57.84 & \textbf{49.80} &
\textbf{64.20} & 92.40 & \textbf{68.04} \\
Skywork-Llama & 52.50 &
Skywork-Llama & 61.15 & 54.35 & 29.90 & 51.80 & 91.00 & 57.64 \\
Lawformer & 68.50 &
Lawformer & 50.95 & 44.92 & 39.00 & 30.40 & 88.60 & 50.77 \\
Legal-BERT & 41.00 &
Legal-BERT & 31.19 & 39.27 & 20.10 & 29.00 & 74.80 & 38.87 \\
\midrule
LexRM-Style & \textbf{81.75} &
LexRM-Element & 77.72 & 57.80 & 44.70 & 59.20 & 92.20 & 66.32 \\
LexRM-Merge & 69.75 &
LexRM-Merge & 75.57 & \textbf{58.27} & 43.80 & 58.80 & 92.40 & 65.77 \\
\bottomrule
\end{tabular*}

\par\vspace{6pt}
\noindent\makebox[\linewidth][c]{\textbf{Chain: LexChain}}
\par\vspace{4pt}

\noindent
\begin{tabular*}{\linewidth}{@{\extracolsep{\fill}}lrrrrrrrr@{}}
\toprule
Method & Plaintiff & Defendant & Dispute & Statute & Liability & Damages & Judgment & Overall \\
\midrule
Five-model average & 96.30 & 87.47 & 18.01 & 21.68 & 22.82 & 23.07 & 24.24 & 45.41 \\
Rubric & 96.00 & 87.45 & 21.20 & 27.25 & 25.75 & 25.85 & 27.82 & 47.80 \\
Skywork-Qwen & 96.75 & 88.60 & 27.00 & 29.95 & 27.95 & \textbf{28.35} & \textbf{32.06} & 50.19 \\
Skywork-Llama & 96.55 & 87.75 & 26.10 & 29.80 & 28.20 & 28.25 & 30.73 & 49.83 \\
Lawformer & 96.20 & 88.15 & 18.20 & 23.65 & 23.05 & 23.15 & 24.53 & 45.93 \\
Legal-BERT & 95.85 & 85.80 & 13.50 & 16.95 & 19.85 & 21.10 & 19.84 & 42.70 \\
\midrule
LexRM-Chain & \textbf{96.80} & \textbf{89.15} & \textbf{27.40} &
\textbf{30.65} & \textbf{28.75} & 28.00 & 31.46 & \textbf{50.46} \\
LexRM-Merge & 96.25 & 88.00 & 19.30 & 23.45 & 24.15 & 26.15 & 26.81 & 46.84 \\
\bottomrule
\end{tabular*}
\end{minipage}
\end{table*}

We evaluate taxonomy-driven rewards at three levels. First, we assess the taxonomy-derived rubric-based rewards on pairwise selection tasks and MoE-based test-time scaling (TTS), where the reward selects the highest-scoring response from five candidates generated by different models, with random selection from the same pool as the baseline. Second, we use rubric-derived preference data to train reward models and evaluate them under the same MoE-based TTS setting.
Third, we assess whether this supervision improves generation quality through DPO on the preference data and reinforcement learning from an SFT checkpoint using the learned reward models.

\subsection{Settings}

\subsubsection{Datasets}

For \textit{Style}, we use CLASE \citep{ma2026clase}, with 4,000 training instances and an official test set of 1,000 instances. Reward scorers are evaluated by pairwise accuracy in selecting the gold response over a model-generated negative; policies are evaluated by the CLASE-Mix score. For \textit{Element}, we draw data from the criminal questions of JEC-QA \citep{zhong2020jecqa} and the civil judgments of LexChain \citep{xie2026lexchain}, using 2,244 preference pairs built from 2,736 instances for training. Evaluation follows the in-domain protocol of Legal$\Delta$ \citep{legaldelta} on its 3,000-instance test set, which reports F1 for statutory-article and charge prediction and accuracy for sentence-length prediction, case analysis, and financial calculation; the overall score is the unweighted mean of the five values. For \textit{Chain}, we use LexChain \citep{xie2026lexchain}, with 9,550 training instances and an official test set of 1,000 instances, and adopt its native LLM-based evaluation, which scores seven aspects of a judgment and reports an overall score. Dataset statistics and the construction of preference data are detailed in Appendix~\ref{b1}. 

\subsubsection{Models}

We use DeepSeek-V4-Flash~\citep{deepseekv4} for all LLM-based components of the rubric evaluators and Qwen3-8B~\citep{qwen3technicalreport} as the backbone for all trained models. To generate candidate responses for preference data construction, we use a pool of five models: Qwen3-4B~\citep{qwen3technicalreport}, Qwen3.5-4B, Qwen3.5-9B~\citep{qwen3.5}, Llama-3.1-8B-Instruct~\citep{dubey2024llama}, and gemma-4-12B-it~\citep{gemmateam2026gemma4}.
To evaluate reward model performance, we compare \textbf{LexRM} against Skywork-Reward-V2-Llama-3.1-8B, Skywork-Reward-V2-Qwen3-8B~\citep{liu2025skywork}, as well as Lawformer~\citep{xiao2021lawformerpretrainedlanguagemodel} and LegalBERT~\citep{chalkidis2020legalbert}.
Training and inference configurations are provided in Appendix~\ref{b2}.

\subsection{Results}

\subsubsection{Rubric and RM Evaluation}

In this section, we evaluate the rubric-based rewards, together with \textbf{LexRM} trained on the preference data we construct, on selecting among candidate responses from multiple models.

As shown in Table~\ref{tab:rm_results}, on all three datasets the rubric-based rewards (\textbf{Rubric}) select better responses than the five-model average or random selection, and the reward models trained on the preference pairs achieve further improvements: 
despite using substantially less training data than the Skywork reward models, \textbf{LexRM-Style} and \textbf{LexRM-Chain} achieve the best overall results on their respective datasets, while \textbf{LexRM-Element} ranks second. 
These results suggest that the reward models successfully internalize the rubric criteria as continuous scoring functions, enabling them to distinguish candidates that the rubrics may score equally.
\textbf{LexRM-Merge} combines the three dimension-specific reward models into one. On Legal$\Delta$, it matches the element expert and obtains the highest charge-prediction score, while on CLASE and LexChain it falls below the respective expert models.

\subsubsection{DPO and RL Evaluation}
\label{dpo and rl}

\begin{table*}[t]
\centering
\caption{Downstream policy performance across the three \textbf{LexReward} dimensions.
SPP-F and CCP report entity-level F1; SLP, CAS and CAC report exact-match accuracy;
Avg. is the mean of these five Element metrics. Bold denotes the best result in each column, including ties.}
\label{tab:dporl_results}

\begin{minipage}{\textwidth}
\footnotesize
\setlength{\tabcolsep}{2pt}
\renewcommand{\arraystretch}{1.12}

\noindent
\begin{tabular*}{\linewidth}{@{\extracolsep{\fill}}lr@{\hspace{10pt}}rrrrrr@{}}
\toprule
\multirow{2}{*}{Method} & \multicolumn{1}{c}{\textbf{Style: CLASE}} &
\multicolumn{6}{c}{\textbf{Element: Legal$\Delta$}} \\
\cmidrule(lr){2-2}\cmidrule(l){3-8}
& CLASE-Mix ($/10$) & SPP-F & CCP & SLP & CAS & CAC & Avg. \\
\midrule
Qwen3-8B    & 3.08 & 76.80 & 51.89 & 45.00 & 35.00 & 80.20 & 57.78 \\
SFT         & 7.32 & 75.45 & 49.85 & 46.80 & 49.20 & 81.80 & 60.62 \\
DPO         & 5.21 & 75.67 & 52.99 & 44.20 & 36.20 & 81.60 & 58.13 \\
\midrule
GRPO (rule) & 8.43 & \textbf{80.45} & 48.67 & 47.60 & 51.80 & 75.80 & 60.86 \\
GRPO (RM)   & \textbf{8.54} & 78.64 & \textbf{54.77} & \textbf{56.70} &
\textbf{53.40} & \textbf{89.80} & \textbf{66.66} \\
\bottomrule
\end{tabular*}

\par\vspace{6pt}
\noindent\makebox[\linewidth][c]{\textbf{Chain: LexChain}}
\par\vspace{4pt}

\noindent
\begin{tabular*}{\linewidth}{@{\extracolsep{\fill}}lrrrrrrrr@{}}
\toprule
Method & Plaintiff & Defendant & Dispute & Statute &
Liability & Damages & Judgment & Overall \\
\midrule
Qwen3-8B    & 97.25 & 88.90 & 34.20 & 42.10 & 34.30 & 28.45 & 26.46 & 53.55 \\
SFT         & 97.95 & \textbf{90.85} & \textbf{39.00} & 39.40 & 35.05 & 35.15 & 36.06 & 55.99 \\
DPO         & 97.50 & 89.05 & 36.20 & \textbf{42.75} & 35.40 & 30.10 & 27.79 & 54.47 \\
\midrule
GRPO (rule) & 97.40 & 89.80 & 36.50 & 37.90 & \textbf{35.55} & 34.00 & 37.64 & 55.29 \\
GRPO (RM)   & \textbf{98.25} & 90.80 & 37.30 & 39.55 & 35.20 &
\textbf{36.70} & \textbf{38.49} & \textbf{56.40} \\
\bottomrule
\end{tabular*}
\end{minipage}
\end{table*}

In this section, we evaluate how rewards derived from the \textbf{LexReward} taxonomy transfer to downstream policies by examining performance after DPO and comparing RL with \textbf{LexRM} against RL with an outcome reward.

As shown in Table~\ref{tab:dporl_results}, DPO improves over the vanilla model across all three dimensions, indicating that the rubric-derived preference data provide useful reward information for policy optimization. Comparing the vanilla model, SFT, and GRPO initialized from SFT further demonstrates the benefits of reinforcement learning: GRPO (RM) achieves the highest score on every dimension, with the largest gain on Element, where it improves all five metrics and raises their average by 6.04 points over SFT. These results show that \textbf{LexRM} provides effective supervision for improving legal generation beyond supervised fine-tuning.


Legal RL is currently driven almost entirely by outcome rewards defined on a verifiable final answer, such as a predicted charge, statute, or monetary amount~\citep{legaldelta, cai2025unilaw, zhang2025legal}. 
Following this practice, we use outcome-based rewards for \textit{Element} and \textit{Chain}, and ROUGE against the reference text for \textit{Style}. 
As shown in Table~\ref{tab:dporl_results}, for \textit{Style}, GRPO (rule) improves performance, but the gain is smaller than that achieved by GRPO (RM), suggesting that lexical overlap provides less effective supervision than the learned reward. 
GRPO (rule) leaves the \textit{Element} and \textit{Chain} averages largely unchanged: its gains are confined to the final answer, whereas the dimensions that ground that answer, such as Legal Basis, stagnate or decline. 
An outcome reward is indifferent to whether the conclusion it credits was reached through a complete and correctly grounded chain. However, rubric-derived rewards are subject to neither restriction. They are equally applicable to open-ended legal scenarios and do not depend on reference labels. 
This shows that the benefit comes from making the reward dimensional: what the taxonomy supplies is supervision on the reasoning that leads to an answer, not a better estimate of the answer itself.




\subsubsection{Dimension Ablation Analysis}

\begin{table*}[t]
\centering
\small
\setlength{\tabcolsep}{3.5pt}
\renewcommand{\arraystretch}{1.0}

\caption{Ablation study of the \textit{Style}, \textit{Element}, and \textit{Chain} rubric dimensions. Bold denotes the best result in each column.}
\label{tab:style_element_chain_rubrics}

\begin{tabular}{@{}ll@{\hspace{0.8em}}lrrrrrr@{}}
\toprule

\multicolumn{2}{c}{\textbf{Style: CLASE}}
&
\multicolumn{7}{c}{\textbf{Element: Legal$\Delta$}}
\\
\cmidrule(r){1-2}
\cmidrule(l){3-9}

Rubric
& Accuracy (\%)
& Rubric
& SPP-F
& CCP
& SLP
& CAS
& CAC
& Avg.
\\
\midrule

Lexical
& 76.50
& Subjects
& 65.70
& 48.76
& 7.60
& 32.80
& \textbf{93.00}
& 49.57
\\

Syntactic
& 77.50
& Facts
& 68.13
& 50.73
& 11.20
& 33.60
& \textbf{93.00}
& 51.33
\\

Overall
& \textbf{79.00}
& Statutes
& 79.52
& \textbf{57.64}
& 30.60
& 39.20
& 92.80
& 59.95
\\

&
&
Disposition
& \textbf{79.54}
& 56.76
& 17.50
& 31.80
& \textbf{93.00}
& 55.72
\\

&
&
Overall
& 79.04
& 57.03
& \textbf{33.10}
& \textbf{39.40}
& 92.80
& \textbf{60.27}
\\

\midrule

\multicolumn{9}{c}{\textbf{Chain: LexChain}}
\\
\cmidrule(lr){1-9}

Rubric &
Plaintiff & Defendant & Dispute & Statute &
Liability & Damages & Judgment & Overall
\\
\midrule

Order
& 95.70
& 87.30
& 19.80
& 24.65
& 23.65
& 23.35
& 25.94
& 46.25
\\

Completeness
& 96.00
& 87.20
& \textbf{21.80}
& \textbf{27.95}
& 24.60
& 24.25
& 27.33
& 47.43
\\

Correctness
& 95.95
& 87.15
& 20.00
& 25.85
& 23.95
& 24.40
& 26.60
& 46.77
\\

Non-redundancy
& \textbf{96.05}
& \textbf{88.15}
& 18.10
& 22.10
& 24.55
& 25.30
& 26.79
& 46.43
\\

Overall
& 96.00
& 87.45
& 21.20
& 27.25
& \textbf{25.75}
& \textbf{25.85}
& \textbf{27.82}
& \textbf{47.80}
\\

\bottomrule
\end{tabular}

\vspace{-2pt}



\end{table*}
In this section, we examine whether the sub-dimensions within each taxonomy dimension carry complementary signal by deriving a reward from each sub-dimension in isolation and comparing it against their aggregation.

As shown in Table~\ref{tab:style_element_chain_rubrics}, aggregation gives the best overall score in all three dimensions of the taxonomy. On individual tasks it is occasionally beaten by a single criterion, but always by a small margin, whereas no single criterion consistently performs best across tasks. Within \textit{Element}, for instance, a rubric using subjects alone never checks which provision is cited, and statute prediction drops well below the aggregate, which is restored by adding statutes; within \textit{Chain}, the aggregate improves over every single criterion, with the gain concentrated in the later stages of the chain, liability, loss and judgment, whose conclusions must be reasoned to rather than read off the case description, while the earlier fact-identification stages are already saturated. This shows that the value of the taxonomy is not that one criterion suffices, but that scoring along several at once keeps a task-irrelevant criterion from deciding the outcome.




\section{Conclusion}

We introduce \textbf{LexReward}, a taxonomy-driven framework that characterizes legal response quality along three complementary dimensions: \textit{Style}, \textit{Element}, and \textit{Chain}. \textbf{LexReward} operationalizes these dimensions as fine-grained scoring rubrics that effectively distinguish responses of different quality. The rubric-derived preference data improve legal response generation through DPO and support the training of \textbf{LexRM}, a family of Chinese legal reward models. 
\textbf{LexRM} improves multi-model response selection and provides effective reinforcement-learning rewards for separately optimizing policies in \textit{Style}, \textit{Element}, and \textit{Chain}.
These results establish taxonomy-driven rewards as effective supervision for improving legal language models.



\section*{AI Use Statement}
In this work, we used generative AI tools for language refinement, partial code implementation, the generation of preference data, and assistance with interpreting results. We did not use these tools to develop theoretical models or conceptual frameworks, formulate mathematical claims, or provide essential components of mathematical proofs; the remaining disclosure categories are not applicable to this work. The authors reviewed and validated all AI-assisted outputs. Specifically, we checked AI-refined text for accuracy, tested AI-generated code and data for correctness, and independently verified AI-assisted interpretations of the results. The authors take full responsibility for the final content of this work, including all text, claims, code, and other artifacts produced with the assistance of generative AI.

\section*{Reproducibility Statement}
The construction of the legal quality taxonomy is described in Section~3, while Section~4 specifies the rubric operationalization, preference data construction, and reward model training objectives. Section~5 presents the datasets, models, and evaluation protocols used in our experiments. Appendix~A provides implementation details for the rubric-based rewards, while Appendix~B documents dataset statistics, preference data construction, and training/inference/evaluation configurations.

\bibliography{iclr2027_conference}

@inproceedings{chalkidis2020legalbert,
  title = {{LEGAL-BERT}: The Muppets straight out of Law School},
  author = {Chalkidis, Ilias and Fergadiotis, Manos and Malakasiotis, Prodromos and Aletras, Nikolaos and Androutsopoulos, Ion},
  booktitle = {Findings of the Association for Computational Linguistics: EMNLP 2020},
  pages = {2898--2904},
  year = {2020},
}

@inproceedings{zhong2020jecqa,
  title = {{JEC-QA}: A Legal-Domain Question Answering Dataset},
  author = {Zhong, Haoxi and Xiao, Chaojun and Tu, Cunchao and Zhang, Tianyang and Liu, Zhiyuan and Sun, Maosong},
  booktitle = {Proceedings of the AAAI Conference on Artificial Intelligence},
  pages = {9701--9708},
  year = {2020},
}

@inproceedings{legalduet,
  title = {{LegalDuet}: Learning Fine-Grained Representations for Legal Judgment Prediction via a Dual-View Contrastive Learning},
  author = {Xu, Buqiang and Dai, Xin and Liu, Zhenghao and Xie, Huiyuan and Yi, Xiaoyuan and Wang, Shuo and Yan, Yukun and Yang, Liner and Gu, Yu and Yu, Ge},
  booktitle = {International Conference on Advanced Data Mining and Applications},
  pages = {337--352},
  year = {2025},
}

@inproceedings{legaldelta,
  title = {{Legal$\Delta$}: Enhancing Legal Reasoning in {LLMs} via Reinforcement Learning with Chain-of-Thought Guided Information Gain},
  author = {Dai, Xin and Xu, Buqiang and Liu, Zhenghao and Yan, Yukun and Xie, Huiyuan and Yi, Xiaoyuan and Wang, Shuo and Yu, Ge},
  booktitle = {Proceedings of the IEEE International Conference on Acoustics, Speech and Signal Processing},
  year = {2026},
  pages = {16912--16916}
}

@inproceedings{legalbench,
  title = {{LegalBench}: A Collaboratively Built Benchmark for Measuring Legal Reasoning in Large Language Models},
  author = {Guha, Neel and Nyarko, Julian and Ho, Daniel E. and R{\'e}, Christopher and Chilton, Adam and Narayana, Aditya and Chohlas-Wood, Alex and Peters, Austin and Waldon, Brandon and Rockmore, Daniel N. and Zambrano, Diego and Talisman, Dmitry and Hoque, Enam and Surani, Faiz and Fagan, Frank and Sarfaty, Galit and Dickinson, Gregory M. and Porat, Haggai and Hegland, Jason and Wu, Jessica and Nudell, Joe and Niklaus, Joel and Nay, John and Choi, Jonathan H. and Tobia, Kevin and Hagan, Margaret and Ma, Megan and Livermore, Michael and Rasumov-Rahe, Nikon and Holzenberger, Nils and Kolt, Noam and Henderson, Peter and Rehaag, Sean and Goel, Sharad and Gao, Shang and Williams, Spencer and Gandhi, Sunny and Zur, Tom and Iyer, Varun and Li, Zehua},
  booktitle = {Advances in Neural Information Processing Systems},
  year = {2023},
  pages= {44123--44279}
}

@inproceedings{lawbench,
  title = {{LawBench}: Benchmarking Legal Knowledge of Large Language Models},
  author = {Fei, Zhiwei and Shen, Xiaoyu and Zhu, Dawei and Zhou, Fengzhe and Han, Zhuo and Huang, Alan and Zhang, Songyang and Chen, Kai and Yin, Zhixin and Shen, Zongwen and Ge, Jidong and Ng, Vincent},
  booktitle = {Proceedings of the 2024 Conference on Empirical Methods in Natural Language Processing},
  pages = {7933--7962},
  year = {2024}
}

@inproceedings{cai2025unilaw,
    title = "Unilaw-R1: A Large Language Model for Legal Reasoning with Reinforcement Learning and Iterative Inference",
    author = "Cai, Hua  and
      Zhao, Shuang  and
      Zhang, Liang  and
      Shen, Xuli  and
      Xu, Qing  and
      Shen, Weilin  and
      Wen, Zihao  and
      Ban, Tianke",
    booktitle = "Proceedings of the 2025 Conference on Empirical Methods in Natural Language Processing",
    year = "2025",
    pages = "18117--18131",
}

@inproceedings{zhang2025legal,
    title = "Legal Mathematical Reasoning with {LLM}s: Procedural Alignment through Two-Stage Reinforcement Learning",
    author = "Zhang, Kepu  and
      Xie, Guofu  and
      Yu, Weijie  and
      Xu, Mingyue  and
      Tang, Xu  and
      Li, Yaxin  and
      Xu, Jun",
    booktitle = "Findings of the Association for Computational Linguistics: EMNLP 2025",
    year = "2025",
    pages = "1586--1598",
}

@inproceedings{ma2026clase,
  title = {CLASE: A Hybrid Method for Chinese Legalese Stylistic Evaluation},
  author = {Ma, Yiran Rex and Ye, Yuxiao and Xie, Huiyuan},
  booktitle = {Proceedings of the Fifteenth Language Resources and Evaluation Conference (LREC 2026)},
  year = {2026},
  pages = {642--653},
}

@article{li2024lexevalcomprehensivechineselegal,
      title={LexEval: A Comprehensive Chinese Legal Benchmark for Evaluating Large Language Models}, 
      author={Haitao Li and You Chen and Qingyao Ai and Yueyue Wu and Ruizhe Zhang and Yiqun Liu},
      year={2024},
      journal={arXiv preprint arXiv:2409.20288}
}

@inproceedings{xie2026lexchain,
  title={LexChain: Modeling Legal Reasoning Chains for Chinese Tort Case Analysis},
  author={Xie, Huiyuan and Li, Chenyang and Zhu, Huining and Zhang, Chubin and Ye, Yuxiao and Liu, Zhenghao and Liu, Zhiyuan},
  booktitle={Proceedings of the AAAI Conference on Artificial Intelligence},
  pages={35913--35921},
  year={2026}
}

@article{qwen3technicalreport,
      title={Qwen3 Technical Report}, 
      author={{Qwen Team}},
      year={2025},
      journal={arXiv preprint arXiv:2505.09388}
}

@article{ouyang2022training,
  title = {Training Language Models to Follow Instructions with Human Feedback},
  author = {Ouyang, Long and Wu, Jeff and Jiang, Xu and Almeida, Diogo
    and Wainwright, Carroll L. and Mishkin, Pamela and Zhang, Chong
    and Agarwal, Sandhini and Slama, Katarina and Ray, Alex
    and Schulman, John and Hilton, Jacob and Kelton, Fraser
    and Miller, Luke and Simens, Maddie and Askell, Amanda
    and Welinder, Peter and Christiano, Paul and Leike, Jan and Lowe, Ryan},
  journal = {Advances in Neural Information Processing Systems},
  year = {2022},
  pages = {27730--27744}
}

@article{wu2023finegrained,
  title = {Fine-Grained Human Feedback Gives Better Rewards for Language Model Training},
  author = {Wu, Zeqiu and Hu, Yushi and Shi, Weijia and Dziri, Nouha
    and Suhr, Alane and Ammanabrolu, Prithviraj and Smith, Noah A.
    and Ostendorf, Mari and Hajishirzi, Hannaneh},
  journal = {Advances in Neural Information Processing Systems},
  year = {2023},
  pages = {59008--59033}
}

@article{wang2024interpretable,
  title = {Interpretable Preferences via Multi-Objective Reward Modeling
    and Mixture-of-Experts},
  author = {Wang, Haoxiang and Xiong, Wei and Xie, Tengyang
    and Zhao, Han and Zhang, Tong},
  journal = {arXiv preprint arXiv:2406.12845},
  year = {2024}
}

@inproceedings{kim2024prometheus,
  title = {Prometheus: Inducing Fine-grained Evaluation Capability
    in Language Models},
  author = {Kim, Seungone and Shin, Jamin and Cho, Yejin and Jang, Joel
    and Longpre, Shayne and Lee, Hwaran and Yun, Sangdoo
    and Shin, Seongjin and Kim, Sungdong and Thorne, James and Seo, Minjoon},
  booktitle = {The Twelfth International Conference on Learning Representations},
  year = {2024},
  pages = {29927--29962}
}

@article{chen2026scprm,
  title = {{SCPRM}: A Schema-aware Cumulative Process Reward Model
    for Knowledge Graph Question Answering},
  author = {Chen, Jiujiu and Liu, Yazheng and Xie, Sihong and Xiong, Hui},
  journal = {arXiv preprint arXiv:2605.02819},
  year = {2026}
}

@inproceedings{yao-etal-2025-elevating,
    title = "Elevating Legal {LLM} Responses: Harnessing Trainable Logical Structures and Semantic Knowledge with Legal Reasoning",
    author = "Yao, Rujing  and
      Wu, Yang  and
      Wang, Chenghao  and
      Xiong, Jingwei  and
      Wang, Fang  and
      Liu, Xiaozhong",
    booktitle = "Proceedings of the 2025 Conference of the Nations of the Americas Chapter of the Association for Computational Linguistics: Human Language Technologies (Volume 1: Long Papers)",
    year = "2025",
    pages = {5630--5642}
}

@article{ilharco2023editing,
  title     = {Editing Models with Task Arithmetic},
  author    = {Ilharco, Gabriel and Ribeiro, Marco Tulio and Wortsman, Mitchell and
               Gururangan, Suchin and Schmidt, Ludwig and Hajishirzi, Hannaneh and
               Farhadi, Ali},
  booktitle = {International Conference on Learning Representations (ICLR)},
  year      = {2023},
  journal = {arXiv preprint arXiv:2212.04089}
}

@misc{qwen3.5,
    title  = {{Qwen3.5}: Towards Native Multimodal Agents},
    author = {{Qwen Team}},
    month  = {February},
    year   = {2026},
    url = {https://qwen.ai/blog?id=qwen3.5}
}

@article{dubey2024llama,
      title={The {L}lama 3 herd of models},
      author={{Llama Team}},
      journal={arXiv preprint arXiv:2407.21783},
      year={2024}
}

@article{gemmateam2026gemma4,
      title={Gemma 4 Technical Report}, 
      author={{Gemma Team}},
      year={2026},
      journal={arXiv preprint arXiv:2607.02770}
}

@article{liu2025skywork,
  title={Skywork-Reward-V2: Scaling Preference Data Curation via Human-AI Synergy},
  author = {Liu, Chris Yuhao and Zeng, Liang and Xiao, Yuzhen and He, Jujie and Liu, Jiacai and Wang, Chaojie and Yan, Rui and Shen, Wei and Zhang, Fuxiang and Xu, Jiacheng and Liu, Yang and Zhou, Yahui},
  journal={arXiv preprint arXiv:2507.01352},
  year={2025}
}

@article{xiao2021lawformerpretrainedlanguagemodel,
      title={Lawformer: A Pre-trained Language Model for Chinese Legal Long Documents}, 
      author={Chaojun Xiao and Xueyu Hu and Zhiyuan Liu and Cunchao Tu and Maosong Sun},
      year={2021},
      journal={arXiv preprint arXiv:2105.03887}
}

@incollection{maley2014language,
  title={The language of the law},
  author={Maley, Yon},
  booktitle={Language and the Law},
  pages={11--50},
  year={2014},
  publisher={Routledge}
}

@book{goodrich1990legal,
  title={Legal discourse: Studies in linguistics, rhetoric and legal analysis},
  author={Goodrich, Peter},
  year={1990},
  publisher={Springer}
}

@article{osbeck2011good,
  title={What is" good legal writing" and why does it matter?},
  author={Osbeck, Mark K},
  journal={Drexel L. Rev.},
  year={2011}
}

@misc{cjo2013,
      author={{China Judgments Online}},
      title={{China Judgments Online}},
      year={2013},
      url = {https://wenshu.court.gov.cn}
}

@article{qi2019openhownet,
  title={Openhownet: An open sememe-based lexical knowledge base},
  author={Qi, Fanchao and Yang, Chenghao and Liu, Zhiyuan and Dong, Qiang and Sun, Maosong and Dong, Zhendong},
  journal={arXiv preprint arXiv:1901.09957},
  year={2019}
}

@misc{jieba,
  author = {Sun, Jian},
  title = {jieba},
  year = {2012},
  url = {https://github.com/fxsjy/jieba}
}

@article{bradley1952rank,
  title={Rank analysis of incomplete block designs: I. the method of paired comparisons},
  author={Bradley, Ralph Allan and Terry, Milton E},
  journal={Biometrika},
  year={1952}
}

@misc{deepseekv4,
  title={{DeepSeek-V4}: Towards Highly Efficient Million-Token Context Intelligence},
  author={DeepSeek-AI},
  year={2026},
  url={https://huggingface.co/deepseek-ai/DeepSeek-V4-Pro/blob/main/DeepSeek_V4.pdf}
}

@article{hurst2024gpt,
      title={{GPT-4o} system card},
      author={OpenAI},
      journal={arXiv preprint arXiv:2410.21276},
      year={2024}
}
\bibliographystyle{iclr2027_conference}

\appendix
\section{Implementation Details of Rubric-Based Rewards}
\subsection{Style}
\label{appendix-style}

\paragraph{Reference and Calibration Corpora.}
The lexical distributions and linguistic reference statistics are constructed from 1,000 authentic Chinese judgments in the CJO ms corpus~\citep{cjo2013}. The normalization scales are estimated separately from the LY fields of 986 CJO ms documents. This separation ensures that each raw metric is normalized to a comparable numerical range.

\paragraph{Word Specificity.}
Legal terms are extracted using a domain-specific terminology lexicon and forward maximum matching (FMM). The maximum and minimum matching lengths are eight and two Chinese characters, respectively. For each term \(t\) in the legal vocabulary \(V_{\mathrm{legal}}\), the response distribution is calculated as

\begin{equation}
Q_y^{\mathrm{legal}}(t)
=
\frac{N_y(t)}
{\sum_{u\in V_{\mathrm{legal}}}N_y(u)},
\label{eq:appendix-legal-distribution}
\end{equation}

where \(N_y(t)\) is the number of occurrences of term \(t\) in \(y\). Terms absent from the response are assigned \(\epsilon=10^{-10}\) before normalization to avoid undefined KL-divergence values. The complete divergence is

\begin{equation}
d_{\mathrm{ws}}(y)
=
\sum_{t\in V_{\mathrm{legal}}}
P^{\mathrm{legal}}(t)
\log
\frac{P^{\mathrm{legal}}(t)}
{Q_y^{\mathrm{legal}}(t)}.
\label{eq:appendix-word-specificity}
\end{equation}

Here, \(P^{\mathrm{legal}}(t)\) is the normalized frequency of \(t\) in the reference corpus. The normalization scale for this dimension is \(s_{\mathrm{ws}}=12.3071\).

\paragraph{Subjective Word Control.}
Subjective expressions are identified using the Chinese HowNet sentiment lexicon\citep{qi2019openhownet} and the same FMM procedure. The reference and response distributions, \(P^{\mathrm{subj}}\) and \(Q_y^{\mathrm{subj}}\), are constructed in the same manner as the legal-term distributions. The resulting KL divergence is normalized using \(s_{\mathrm{sw}}=12.6957\).

\paragraph{Sentence Cohesion.}
The response is tokenized and POS-tagged using jieba.posseg~\citep{jieba}. Tokens with the POS tag \texttt{c} are treated as conjunctions. The reference conjunction rate is

\begin{equation}
\rho_{\mathrm{conj}}^{*}=0.0191.
\label{eq:appendix-reference-conjunction}
\end{equation}

Here, \(\rho_{\mathrm{conj}}^{*}\) is the mean conjunction-token proportion estimated from the reference documents. The absolute deviation from this value is normalized using \(s_{\mathrm{coh}}=0.011573\).

Pronoun frequency is not included because preliminary experiments showed that it provided little discrimination between authentic and generated responses.

\paragraph{Sentence Structure.}
Sentences are segmented using Chinese punctuation delimiters and newline characters. Sentence length is measured in Chinese characters. The reference statistics are

\begin{equation}
\mu^{*}=64.7,
\qquad
\sigma^{*}=45.3,
\label{eq:appendix-reference-sentence-length}
\end{equation}

where \(\mu^{*}\) and \(\sigma^{*}\) are the reference mean and standard deviation of sentence lengths. The Euclidean distance between the response and reference statistics is normalized using \(s_{\mathrm{str}}=26.0567\).

\begin{table}[t]
\centering
\label{tab:style-normalization-scales}
\begin{tabular}{lll}
\toprule
Dimension & Raw deviation & Scale \(s_k\) \\
\midrule
WS  & KL divergence              & 12.307  \\
SW & KL divergence              & 12.696  \\
COH  & Conjunction-rate deviation & 0.012 \\
STR  & Sentence-statistic distance & 26.057 \\
COL  & \(n\)-gram coverage distance & 0.282 \\
\bottomrule
\end{tabular}
\caption{Normalization scales for the Style reward. WS: Word Specificity; SW: Subjective-Word Control; COH: Sentence Cohesion; STR: Sentence Structure; COL: Collocation Correctness.}
\end{table}

\paragraph{Collocation Correctness.}
Both the reference documents and generated responses are tokenized using jieba, with punctuation tokens removed before \(n\)-gram extraction. A reference \(n\)-gram is retained only if it occurs in at least ten reference documents.

For each \(n\in\{3,4,5,6\}\), let \(G_n(y)\) denote the set of \(n\)-grams extracted from response \(y\), and let \(G_n^{*}\) denote the retained reference set. The coverage statistic is

\begin{equation}
c_n(y)
=
\frac{|G_n(y)\cap G_n^{*}|}
{|G_n(y)|},
\qquad
n\in\{3,4,5,6\}.
\label{eq:appendix-ngram-coverage}
\end{equation}

Here, \(|G_n(y)\cap G_n^{*}|\) is the number of response \(n\)-grams found in the reference set, while \(|G_n(y)|\) is the total number of response \(n\)-grams.

The collocation deviation is calculated as

\begin{equation}
d_{\mathrm{col}}(y)
=
\sqrt{
\sum_{n=3}^{6}
\left(
c_n(y)-\bar{c}_n^{*}
\right)^2
},
\label{eq:appendix-collocation-distance}
\end{equation}

where \(\bar{c}_n^{*}\) is the mean reference coverage for order \(n\). We use \(3\)-grams through \(6\)-grams in the final metric. Bigrams are excluded because their high frequency and short length result in many generic or non-legal combinations, reducing their ability to characterize professional legal collocations.

\paragraph{Normalization Parameters.}
The complete set of normalization scales is summarized below:

For every dimension, the normalized reward is \(r_k(y)=-d_k(y)/s_k\). The five normalized rewards are then combined using an equal-weight arithmetic mean.

\subsection{Element}
\label{appendix-element}

\begin{figure*}[t]
  \centering
  \includegraphics[width=\textwidth]{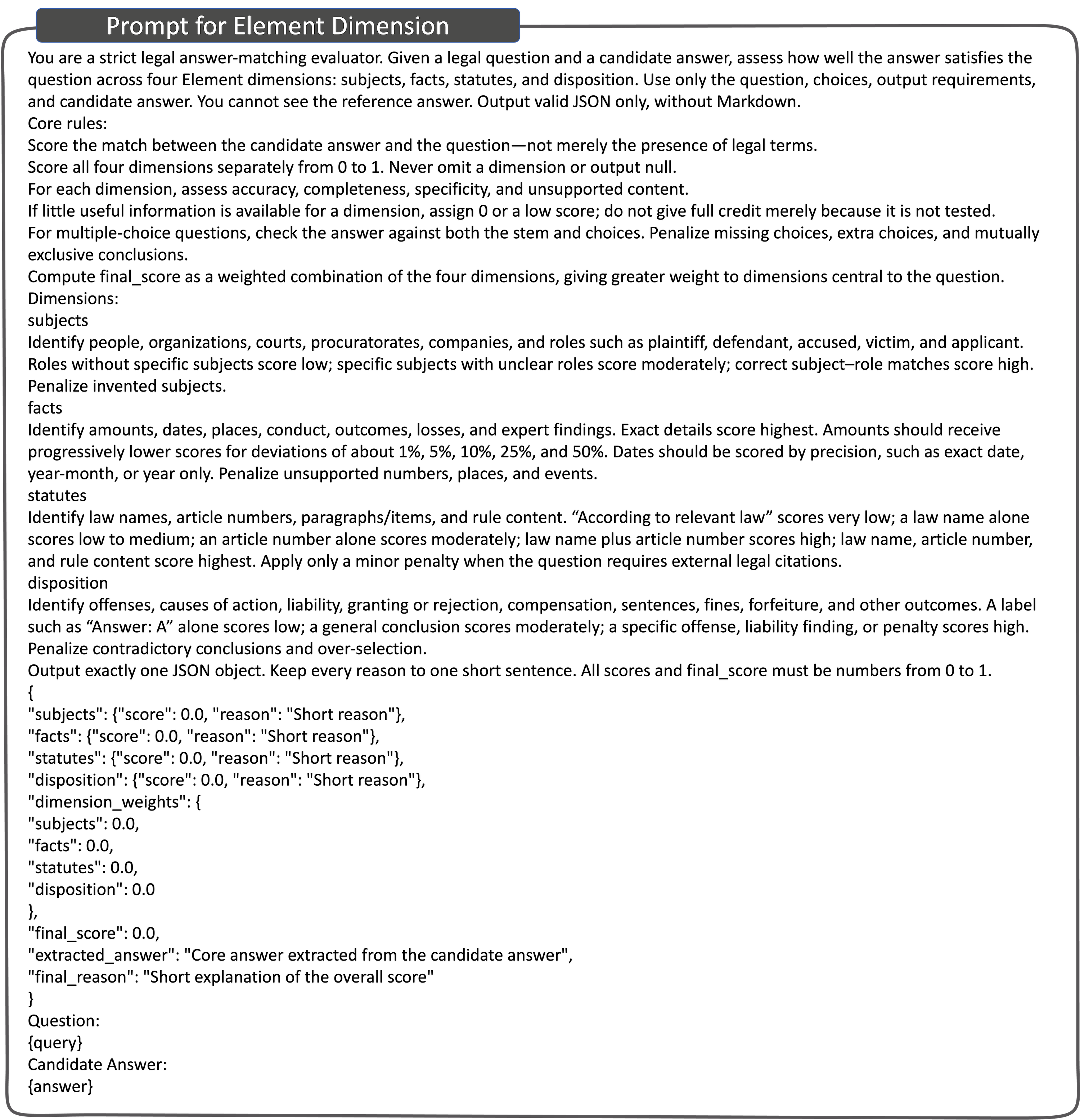}
  \caption{English translation of the rubric-guided prompt used for \textit{Element} evaluation. The original experimental prompt was written in Chinese, while its evaluation criteria, scoring procedure, and output structure are preserved in the translation.}
\label{fig:element-prompt}
\end{figure*}

Fig.~\ref{fig:element-prompt} presents the prompt template used by the LLM-based \textit{Element} evaluator. The prompt provides the task context, candidate response, and element-specific rubric criteria, and requires the judge to produce a structured assessment of the relevant legal elements. For readability, the prompt shown in the figure has been translated into English, and the original prompt used in all experiments was written in Chinese.

\subsection{Chain}
\label{appendix-chain}

The \textit{Chain} reward uses rule-based scoring when task-specific scoring logic is available. We evaluate four criteria: \textit{Order}, which detects violations of required step precedence; \textit{Completeness}, which measures coverage of necessary reasoning steps; \textit{Correctness}, which checks compliance with step-level reasoning requirements; and \textit{Non-redundancy}, which identifies repeated or conflicting steps. Detected omissions or violations are mapped to discrete criterion-level rewards, which are averaged to obtain the final Chain reward.

\begin{figure*}[t]
  \centering
  \includegraphics[width=\textwidth]{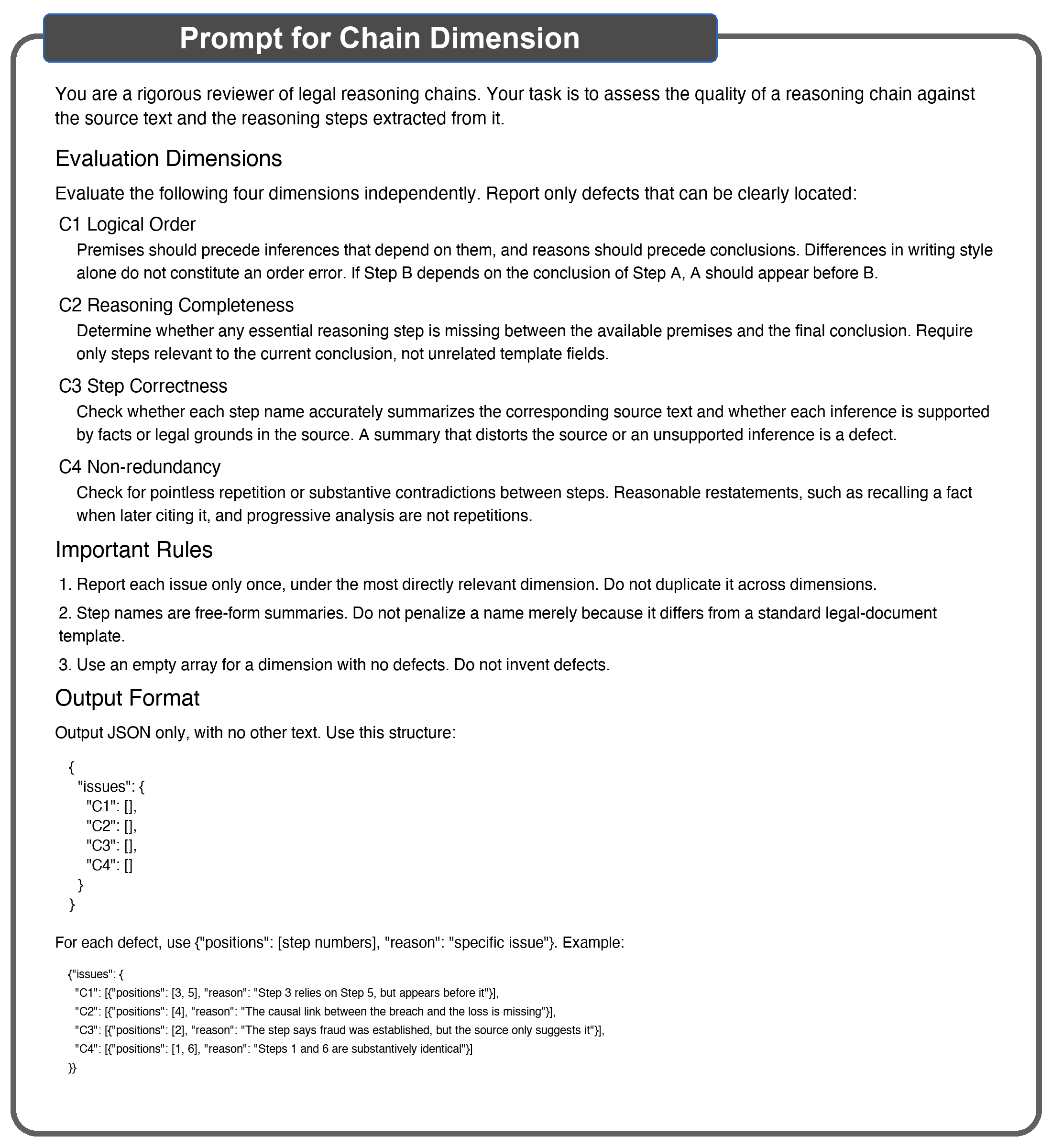}
  \caption{English translation of the reasoning-step extraction prompt used for \textit{Chain} evaluation when predefined reasoning steps are unavailable. The LLM converts free-form reasoning into an ordered sequence of labeled steps, which is subsequently scored by rule-based rubrics. The original experimental prompt was written in Chinese.}
  \label{fig:chain-prompt}
\end{figure*}

When task-specific scoring logic is unavailable, we use an LLM judge to score the response along the same four criteria using the prompt in Fig.~\ref{fig:chain-prompt}. The criterion-level scores are averaged to obtain the final reward. The displayed prompt is translated into English; the original experimental prompt was written in Chinese.

\section{Experimental Settings}
\label{app:para_details}
\subsection{Datasets}
\label{b1}
For Style, we use the official CLASE test set of 1,000 response pairs and construct preference data from all 4,000 training queries using our five-model generation procedure. We reserve 2,000 training instances as the reference corpus for computing the objective component of CLASE-Mix and split the remaining 2,000 queries equally between SFT and GRPO. No model is trained on authentic judicial texts: SFT targets and preferred responses in preference pairs are model-generated candidates selected by their rubric scores, while GRPO uses the learned reward model to score online generations.

For Element, we use all 2,736 JEC-QA and LexChain instances for preference construction, yielding 2,244 preference pairs, and designate separate subsets for SFT and GRPO. Evaluation uses the 3,000 instances in the Legal$\Delta$ in-domain test set, comprising 500 instances each for statutory-article prediction, charge prediction, case analysis, and financial calculation, and 1,000 for sentence-length prediction. 

For Chain, we allocate 1,000 of the 9,550 LexChain training instances to SFT, 1,000 to GRPO, and 4,000 to pairwise preference training, and evaluate on the official test set of 1,000 instances.

Across all three dimensions, we construct preference data by sampling one response per query from each model in the pool, scoring the five candidates with the corresponding rubric, and forming pairs from responses with different scores. Queries for which all candidates receive identical scores are discarded.

\subsection{Hyper-parameters}
\label{b2}
\textbf{Inference}. For LLM-based rubric evaluation we use the DeepSeek-V4-Flash\citep{deepseekv4} API with the default reasoning effort and a maximum input length of 8,192 tokens; all inputs fit within this limit, requiring no truncation. Candidate responses are sampled with temperature 0.8 and top-$p$ 0.95, one response per model.


\textbf{Evaluation}. CLASE-Mix combines objective and subjective scores with equal weights. The objective component compares textual features of generated responses with those of authentic judicial texts, while the subjective component uses GPT-4o-mini to score responses with the benchmark-provided prompt. For LexChain, we use GPT-4o-2024-05-13~\citep{hurst2024gpt} to score each evaluation dimension following the benchmark protocol. For Legal$\Delta$, we report F1 scores for statutory-article and charge prediction, and accuracy for case analysis, financial calculation, and sentence-length prediction. We follow the evaluation procedures specified in the respective benchmark papers; further details can be found therein.

\textbf{Training}. Reward models are trained with the Bradley--Terry objective~\citep{bradley1952rank} for two epochs using DeepSpeed ZeRO-3, a learning rate of $1\times10^{-7}$ and a batch size of 32; DPO uses the same configuration. We train policies separately for Style, Element, and Chain. For each dimension, we first perform SFT on 1,000 queries paired with their highest-scoring candidate responses under the corresponding rubric, then apply GRPO to another 1,000 queries. Both GRPO variants start from the same dimension-specific SFT checkpoint: GRPO (RM) uses the corresponding dimension-specific LexRM as its reward function, whereas GRPO (rule) uses the baseline reward described in Section~\ref{dpo and rl}. Each dimension block in Table~\ref{tab:dporl_results} reports results from the respective policies. DPO is initialized directly from Qwen3-8B. SFT, DPO and GRPO all use low-rank adaptation (LoRA).

\section{Limitations}

Our taxonomy is grounded in Chinese legal materials input from the Chinese legal context, and our evaluation is therefore limited to Chinese-language datasets. Its applicability to other languages and legal systems remains untested. Future work will extend the framework to these settings and examine how the taxonomy and its reward criteria should be adapted.

This work primarily studies dimension-specific rewards. Combining multiple dimensions into a unified rubric reward or reward model remains an open challenge, including the choice of aggregation weights, the composition of preference data, and the training strategy. Developing and systematically evaluating such integration methods is a central direction for future work.

\end{document}